\documentclass[letterpaper, 10 pt, conference]{ieeeconf}  % Comment this line out if you nee    d a4paper

\IEEEoverridecommandlockouts                              % This command is only needed if 
\usepackage{graphicx} % for pdf, bitmapped graphics files
\usepackage{amsmath} % assumes amsmath package installed
\usepackage{amssymb}  % assumes amsmath package installed
\usepackage{bm}
\usepackage{color,soul}
\usepackage{booktabs}
\usepackage{comment}
\usepackage{Shorthands}
\usepackage[linesnumbered,ruled,vlined]{algorithm2e}
\usepackage{nicematrix}
\SetKwInput{KwInput}{Input}               
\SetKwInput{KwOutput}{Output}  

\DeclareMathAlphabet\mathbfcal{OMS}{cmsy}{b}{n}
\newcommand\inv[1]{#1\raisebox{1.15ex}{$\scriptscriptstyle-\!1$}}
\usepackage{subfig}
\makeatletter
\let\NAT@parse\undefined
\makeatother
\usepackage[hidelinks]{hyperref}

\title{\LARGE \bf
Flying Quadrotors in Tight Formations\\using Learning-based Model Predictive Control
}

\author{Kong Yao Chee$^{*}$, Pei-An Hsieh$^{*}$, George J. Pappas, M. Ani Hsieh% <-this % stops a space
\thanks{
We gratefully acknowledge the support of Office of Naval Research (ONR) Award No. N00014-22-1-2157 and DSO National Laboratories, 12 Science Park Drive, Singapore 118225.}
\thanks{$^{*}$: Equal contribution. The authors are with the GRASP Laboratory, University of Pennsylvania, Philadelphia, PA 19104, USA.
        {\tt\footnotesize \{ckongyao,\, pahsieh,\,pappasg,\,m.hsieh\}@seas.upenn.edu}}% <-this % stops a space
}

\begin{document}

\maketitle
\thispagestyle{empty}
\pagestyle{empty}

%%%%%%%%%%%%%%%%%%%%%%%%%%%%%%%%%%%%%%%%%%%%%%%%%%%%%%%%%%%%%%%%%%%%%%%%%%%%%%%%
\begin{abstract}
Flying quadrotors in tight formations is a challenging problem. It is known that in the near-field airflow of a quadrotor, the aerodynamic effects induced by the propellers are complex and difficult to characterize. Although machine learning tools can potentially be used to derive models that capture these effects, these data-driven approaches can be sample inefficient and the resulting models often do not generalize as well as their first-principles counterparts. In this work, we propose a framework that combines the benefits of first-principles modeling and data-driven approaches to construct an accurate and sample efficient representation of the complex aerodynamic effects resulting from quadrotors flying in formation. The data-driven component within our model is lightweight, making it amenable for optimization-based control design. Through simulations and physical experiments, we show that incorporating the model into a novel learning-based nonlinear model predictive control (MPC) framework results in substantial performance improvements in terms of trajectory tracking and disturbance rejection. In particular, our framework significantly outperforms nominal MPC in physical experiments, achieving a 40.1\% improvement in the average trajectory tracking errors and a 57.5\% reduction in the maximum vertical separation errors. Our framework also achieves exceptional sample efficiency, using only a total of 46 seconds of flight data for training across both simulations and physical experiments. Furthermore, with our proposed framework, the quadrotors achieve an exceptionally tight formation, flying with an average separation of less than 1.5 body lengths throughout the flight. A video illustrating our framework and physical experiments is given \href{https://youtu.be/Hv-0JiVoJGo}{\textcolor{blue}{here}}.
\end{abstract}

%%%%%%%%%%%%%%%%%%%%%%%%%%%%%%%%%%%%%%%%%%%%%%%%%%%%%%%%%%%%%%%%%%%%%%%%%%%%%%%%
% \vspace{-0.4cm}
\section{INTRODUCTION} \label{sec:intro}
% \vspace{-0.15cm}
Quadrotor teams have emerged as leading contenders for practical tasks such as surveillance \cite{acevedo2013cooperative}, fighting forest fires \cite{ghamry2016cooperative}, payload transportation \cite{jin2022adaptive}, and urban air transport \cite{ventura2022high}. However, the difficulty of close proximity flights impedes critical subtasks such as docking \cite{shankar2023docking}, formation flight \cite{jasim2017robust}, and other cooperative maneuvers \cite{grli2023nonlinear}. In particular, the complex aerodynamic effects that manifest within a quadrotor team \cite{gielis2023modeling} makes it challenging for them to fly in tight formations. Hence, the ability to characterize and compensate for these aerodynamic effects would enable them to achieve tighter formations and accomplish these tasks more effectively.

\textit{Related works:} One possible approach to characterizing these aerodynamic effects is to derive a first-principles model based on physics and empirical data \cite{jain2019modeling, xu2024omnidrones, bauersfeld2024robotics}. This model can then be used for controller design to counteract these effects or disturbances. A key limitation of this approach is that these models are often overly simplified and do not provide accurate predictions of the velocity flow field and representations of the aerodynamic effects. A recent study in \cite{kiran2024downwash} shows that the flow field near a quadrotor is highly complex and nonlinear, which makes the modeling of these aerodynamic effects using first principles particularly challenging. An alternative approach is to first collect data and extract a data-driven representation \cite{gielis2023modeling, smith2023so}. These data-driven models, however, are significantly less sample efficient compared to first-principles models. Moreover, this approach may require an onerous data collection procedure, such as those described in \cite{smith2023so, li2023nonlinear}. Instead, we leverage a state-of-the-art deep learning tool, knowledge-based neural ordinary differential equations (KNODE) \cite{jiahao2021knowledge}. This allows us to construct a knowledge-based, data-driven representation that only requires a small amount of data to train. More importantly, this representation captures the complex aerodynamic effects with unprecedented accuracy.

\begin{figure}
    \centering
    {\includegraphics[scale=0.475, trim = 0cm 0.4cm 0cm -0.3cm]{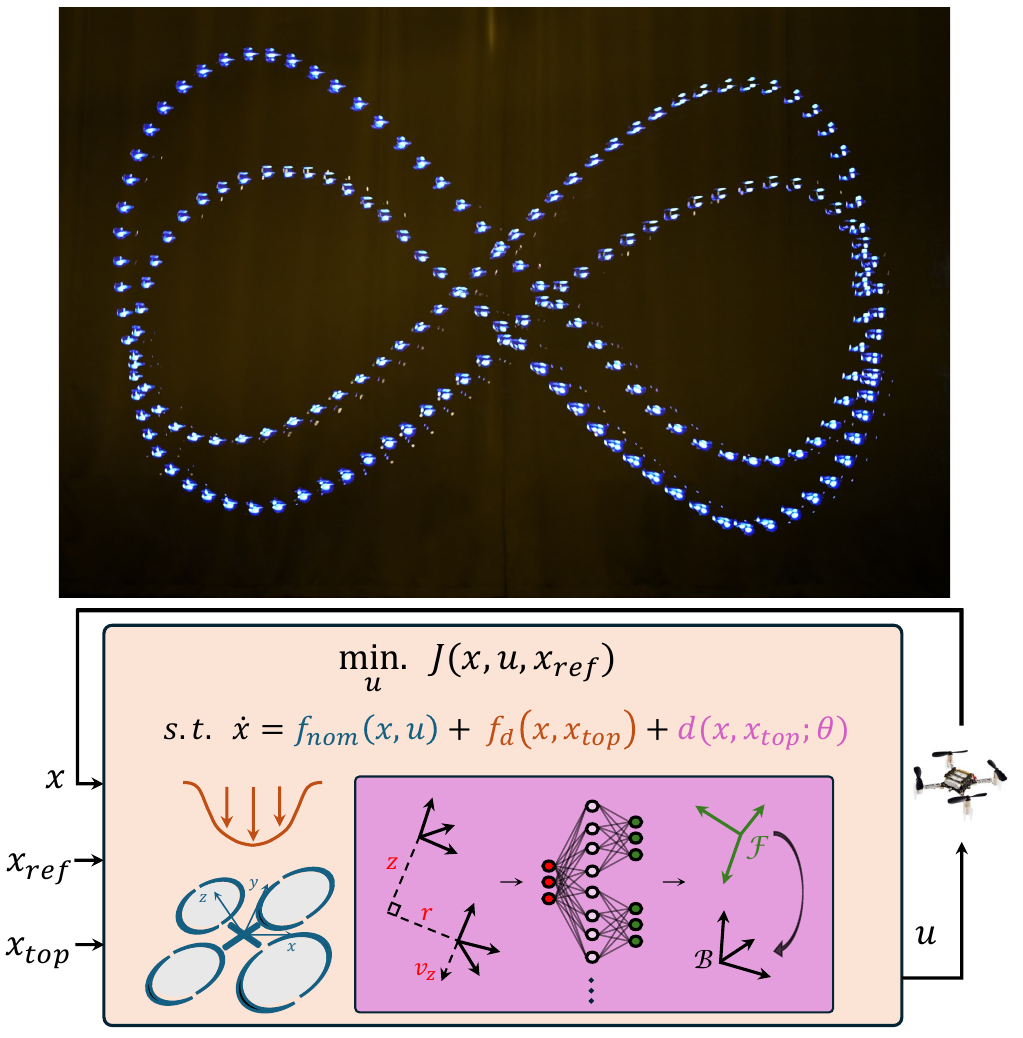}}
    \caption{\small\textbf{KNODE-DW MPC:} \textit{Top}: A composite photo depicting two Crazyflie (CF) quadrotors flying in lemniscate trajectories, under a stacked formation with a commanded separation of 2 body lengths. This is achieved with our proposed framework. The frames used to create this photo are extracted from a video taken during a physical experiment. \textit{Bottom}: Schematic of our framework. \textit{CF image in schematic}: \cite{Bitcraze}.}
    \label{fig:cover_image}
\end{figure}

To enable the quadrotors to fly in close proximity, the derived models are often deployed in reactive controllers, \textit{e.g.}, \cite{smith2023so, shi2020neural}. These controllers consider the aerodynamic effects or disturbances at the current time step, but neglect potential disturbances that may occur within the next few time steps. On the other hand, model predictive control (MPC) offers an opportunity to anticipate future aerodynamic disturbances, through the consideration of a prediction horizon. In \cite{li2023nonlinear}, a learned model is incorporated into an MPC framework, allowing two quadrotors to fly close to each other. However, the tests considered in \cite{li2023nonlinear} are only limited to short-duration traversals between two quadrotors, and are conducted at one particular speed and separation distance. Hence, it is unclear if the framework can generalize to different flight conditions and maneuvers. In contrast, our proposed framework achieves high accuracy across a wide range of test cases. These include a particularly challenging \emph{stacked} formation, where a quadrotor constantly flies in the wake of another quadrotor. The aerodynamic effects in this case are highly transient, making it extremely challenging for the quadrotors to achieve close proximity flight.

\textit{Contributions:} The main contributions of this work are three-fold. First, we introduce a novel knowledge-based, data-driven model that accurately characterizes the complex aerodynamic disturbances within a quadrotor team. We combine analytical models of the quadrotor dynamics and aerodynamic disturbances, which we refer to as \emph{knowledge} of the system, with a data-driven component. %In addition to a data-driven component, we consider analytical models of the quadrotor dynamics and aerodynamic disturbances, which we refer to as \emph{knowledge} of the system. 
Second, we incorporate the combined model into a nonlinear MPC framework, as depicted in the schematic in Fig \ref{fig:cover_image}. We ascertain the closed-loop performance through extensive simulations. Notably, we show that our learning-based MPC framework achieves performance comparable to an omniscient MPC framework, where the model in the controller is identical to the true dynamics. Third, we demonstrate through physical experiments that our framework achieves accurate and consistent closed-loop performance across numerous test cases. In particular, our framework generalizes exceptionally well beyond the training dataset. With our framework, the quadrotors fly in an incredibly tight formation, where they maintain an average separation distance of less than 1.5 body lengths throughout the entire flight. To the best of the authors' knowledge, this is a feat never achieved before for a quadrotor team of this size and weight class. 

% \vspace{-0.1cm}
\section{PRELIMINARIES} \label{sec:prelim}
% \vspace{-0.1cm}
\subsection{Quadrotor Dynamics and Aerodynamic Disturbances} \label{subsec:dynamics}
% \vspace{-0.1cm}
Consider a quadrotor system with the dynamics \cite{chee2023enhancing},
% \vspace{-0.1cm}
\begin{equation}\label{eq:eoms}
\begin{split}
    \dot{p} = v,\;\;\,\quad\qquad\qquad
    \dot{v} &= -g + (1/m)R \tau,\\
    J\dot{\omega} = \alpha - \omega \times J \omega,\qquad
    \dot{q} &= \Omega \, (q/2),
\end{split}
% \vspace{-0.2cm}
\end{equation}
where $p,\,v,\,\omega :=[\omega_x,\omega_y,\omega_z]^{\top}\in \setR^3$ and $q \in \setR^4$ are the position, velocity, angular rate and quaternions representing its dynamics and kinematics. The vectors $\tau := [0\;\,0\;\,\eta]^{\top} \in \mathbb{R}^3$ and $\alpha \in \mathbb{R}^3$ consist of the thrust $\eta$ and torques generated by the motors of the quadrotor. The matrix $R \in \mathbb{R}^{3\times3}$ is the rotation matrix from the body $(\calB)$ to the world $(\calW)$ reference frame. The world frame follows the East-North-Up convention where the $z$ axis is pointing upwards. The matrix $\Omega \in \setR^{4\times 4}$ is defined as 
% \vspace{-0.1cm}
\begin{equation}
\Omega := \begin{bmatrix}0 & -\omega_x & -\omega_y & -\omega_z\\\omega_x& 0 & \omega_z & -\omega_y\\\omega_y & -\omega_z & 0 & \omega_x\\ \omega_z & \omega_y & -\omega_x & 0\end{bmatrix}. 
% \vspace{-0.1cm}
\end{equation}
The mass and inertia matrix of the quadrotor are denoted by $m$ and $J\in \setR^{3\times3}$ respectively. By defining the state and control inputs as $x := [p^{\top}\;v^{\top}\;q^{\top}\;\omega^{\top}]^{\top}$ and $u := [\eta\; \alpha^{\top}]^{\top}$, the equations of motion \eqref{eq:eoms} can be written in a compact form,
% \vspace{-0.3cm}
\begin{equation} \label{eq:quad_dyn}
    \dot{x} = f_{nom}(x,u),
% \vspace{-0.1cm}
\end{equation}
which we refer to as the nominal dynamics model.

Consider a scenario where two quadrotors fly in a tight formation where they must maintain a fixed vertical separation distance. We refer to them as the bottom and top quadrotors. The positions of the quadrotors are denoted by $p \in \mathbb{R}^{3}$ and $p_{top} \in \mathbb{R}^{3}$, and the position difference is given by $\Delta p := p - p_{top}$. By denoting the unit vector of the body frame of the top quadrotor in the $z$ direction as $e_3'$, we define a flow field reference frame $\calF$ with the unit vectors $\{e_1, e_2, e_3\}$ given as
% \vspace{-0.1cm}
\begin{equation}
    e_1 = e_2 \times e_3, \quad 
    e_2 = \frac{e_3 \times \Delta p}{||e_3 \times \Delta p||_2} , \quad 
    e_3 = -e_3',
    % R_{\calW}^{\calF} &:= \begin{bmatrix}
    % -e_3' \times e_2 & e_2 & -e_3' 
    % \end{bmatrix}^{\top},
\end{equation}
and the rotation matrix $R^{\calF}_{\calW} \in \mathbb{R}^{3\times 3}$ rotates a vector from $\mathcal{W}$ to $\calF$.   

To characterize the disturbances on the bottom quadrotor, we consider a velocity flow model \cite{bauersfeld2024robotics} where the relative velocity of the airflow towards a point $p_f := [x_f\;y_f\;z_f]^{\top} \in \mathbb{R}^3$ on the bottom quadrotor in the frame $\calF$ is given as
% \vspace{-0.1cm}
\begin{equation}
\label{eq:aero_dis}
V\left(z, r\right) = \frac{U_H \frac{Bd}{\tilde{z} - z_0}
}{\left( 1 + \left(\sqrt{2} - 1\right) \left(\frac{ \tilde{r}}{S\left(\tilde{z} - z_0\right)}
\right)^2 \right)^2}
 -v_z^{\calF}, 
\end{equation}
where $S \in \mathbb{R}$ is the spreading factor and the velocity $v^{\calF} = R^{\calF}_{\calW} \, v := [v_x^{\calF}\; v_y^{\calF}\;v_z^{\calF}]^{\top} \in \mathbb{R}^3$. The radial and vertical separations between $p_f$ and the top quadrotor in the frame $\calF$ are denoted as $r$ and $z$ respectively, where $r := {\sqrt{x_f^2 + y_f^2}}$ and $z := z_f$. The normalized separations are given as $\tilde{r} := r/\lambda$ and $\tilde{z} := z / \lambda$. For more details on the flow model, readers are referred to \cite{bauersfeld2024robotics}.

Using \eqref{eq:aero_dis}, the disturbance forces $\tau_{d} \in \mathbb{R}^3$ and torques $\alpha_{d} \in \mathbb{R}^3$ are computed as in \cite{jain2019modeling},
\begin{subequations} \label{eq:aero_force_torque}
% \vspace{-0.1cm}
\begin{align}
\tau_{d}\left(x, x_{top}\right) &= \int_{A_b} e_3\, C_D \, \zeta \, dA, \\
\alpha_{d}\left(x, x_{top}\right) &= \int_{A_b} \Big( p_f - R^{\calF}_{\calW} \Delta p \Big) \times  e_3\, C_D \, \zeta \, dA,
\end{align}
\end{subequations}
where the dynamic pressure is denoted by $\zeta := \frac{1}{2} \rho V(z, r)^2 \in \mathbb{R}$, $C_D \in \mathbb{R}$ is the drag coefficient, and the vector $x_{top} \in \mathbb{R}^{13}$ denotes the state of the top quadrotor. The surface area of the bottom quadrotor is a circle, \textit{i.e.}, $A_b := \pi(\lambda/2)^2$. These forces and torques \eqref{eq:aero_force_torque} are succinctly expressed as
% \vspace{-0.2cm}
\begin{equation} \label{eq:f_d}
\begin{split}
    \dot{x}_{dw} &= \left[ 0_{1\times 3} \;\;\; \tfrac{\left(R_{\calW}^{\calF\, \top}\tau_{d}\right)^{\top}}{m}  \;\;\; 0_{1 \times 4} \;\;\;  {\inv{J} \left(R_{\calB}^{\calF \, \top} \alpha_{d}\right)^{\top}} \right]^{\top}\\
    &:= f_d\left(x, x_{top}\right).
\end{split} 
\end{equation}

It is important to note that the model \eqref{eq:f_d} is unlikely to be highly accurate in practice, due to the complexity of the near-field airflow of the quadrotor, as discussed in \cite{kiran2024downwash}. This is particularly evident when the quadrotors are in a tight formation, \textit{i.e.} when the vertical separation within the quadrotors is less than 2 body lengths. In these conditions, the flow model \eqref{eq:aero_dis} is likely to provide accurate predictions only in the far-field airflow below the top quadrotor \cite{bauersfeld2024robotics}. Given that it is challenging to derive a model that accurately characterizes the near-field airflow, our approach is to leverage this simplified model \eqref{eq:f_d} as part of our prior knowledge within the overall dynamics model, and use a data-driven component to learn the complex dynamics of the aerodynamic disturbances, described in Section \ref{subsec:model_train}.

% \vspace{-0.2cm}
\subsection{Learning-based MPC} \label{subsec:learning_mpc}
% \vspace{-0.1cm}
To control the quadrotor system, we consider the following finite horizon optimal control problem \cite{borrelli2017predictive},
\begin{subequations} \label{eq:ftocp}
% \vspace{-0.23cm}
\begin{align}
    \underset{\{x_i\},\{u_i\}}{\textnormal{min.}}\quad\; & \sum_{i=0}^{N-1} \left|\left|x_i- x_{r,i}(k)\right|\right|^2_Q + ||u_i||_R^2 \\
    & + \left|\left|x_N-x_{r,N}(k)\right|\right|^2_P \label{eq:term_mpc_cost}\\
    \text{s.t.}\quad\; &x_{i+1} = {\hat{f}(x_i, u_i;\, \theta^{\star})}, \quad i=0,\dots,N-1, \label{eq:dynamics_const}\\ 
    \; &x_i \in \mathcal{X}, \quad u_i \in \mathcal{U}, \quad i=0,\dots,N-1,\\
    \; &x_N \in \mathcal{X}_f,\quad x_0 = x(k), \label{eq:term_constraint}
\end{align}
\end{subequations}
where $N \in \mathbb{N}_+$ denotes the prediction horizon and the notation $||x||_A$ denotes $x^{\top} A x$. The matrices $Q$, $R$ and $P$ are the cost matrices for the state, control and terminal state costs respectively. The sets $\mathcal{X}$, $\mathcal{X}_f$ and $\mathcal{U}$ define the state, terminal state and control input constraints. The vector $x(k)$ is the state measurement at the $k^{\text{th}}$ time step. The sequence $\{x_{r,i}(k)\}_{i=0}^N$ denotes the reference state trajectories to be tracked by the system. At each time step, the problem \eqref{eq:ftocp} is solved in a receding horizon manner to obtain a sequence of optimal control inputs $\left\{u_i^{\star}\right\}_{i=0}^{N-1}$, and the first element of the sequence $u_0^{\star}$ is applied as the control action to the system. 

The key difference between \eqref{eq:ftocp} and that in a nominal MPC framework lies in the dynamics constraints \eqref{eq:dynamics_const}. In nominal MPC, a nominal model \eqref{eq:quad_dyn} is used as the dynamics constraints. However, due to the presence of unmodeled dynamics, this nominal model may not be sufficiently accurate in representing the true system dynamics, which can cause a degradation in the closed-loop performance. To alleviate this issue, a learning-based framework is applied. A model with trained parameters $\theta^{\star}$ is used to characterize the full dynamics, which includes the unmodeled dynamics. This model is then applied to the constraints \eqref{eq:dynamics_const}. It has been shown in prior work that a learning-based MPC framework provides significant performance improvements over nominal MPC \cite{chee2022knode, chee2022learning}. In this work, we apply a similar methodology, but with some enhancements to accommodate the aerodynamic disturbance model \eqref{eq:f_d} and the predictions of the state of the top quadrotor $x_{top}$.

% \vspace{-0.15cm}
\section{MODEL AND CONTROL DESIGN} \label{sec:model_control}
% \vspace{-0.15cm}
\subsection{Model Formulation and Training} \label{subsec:model_train}
% \vspace{-0.15cm}
For the dynamics model, we consider a state-of-the-art deep learning tool, knowledge-based neural ODEs (KNODE) \cite{jiahao2021knowledge, chen2018neuralode}, to incorporate knowledge of the quadrotor dynamics and aerodynamic disturbances during training and within the controller. First, we parameterize the residual aerodynamic effects that are not captured by knowledge using a neural ODE $d(x,x_{top}; \,\theta)$. This neural ODE is constructed as a $\Upsilon$-branch, L-layer feedforward neural network,  
\begin{subequations} \label{eq:neural_ode}
% \vspace{-0.2cm}
\begin{align}
    h_{0, \upsilon} &= \left[z,\; r,\; v_z^{\calF}\right] ^{\top},\quad {\upsilon} = 1,\, \dots,\, {\Upsilon}, \label{eq:h0} \\
    % h_{l+1, \upsilon} &= \sigma\left(W_{l, \upsilon} h_{l, \upsilon} + b_{l, \upsilon}\right), \\
    h_{L, \upsilon} &= \text{MLP}_{\upsilon}(h_{0,\upsilon}),\quad {\upsilon} = 1,\, \dots,\, {\Upsilon}, \label{eq:mlp_output} \\
    % h_{L} &= [h_{L, 0}^{\top}, \dots, h_{L, \Upsilon}^{\top}]^{\top},\\
    d(x,x_{top}; \theta) &:= MH\, [h_{L, 1}^{\top},\; \dots,\; h_{L, \Upsilon}^{\top}]^{\top},
\end{align}
\end{subequations}
where $H \in \mathbb{R}^{13 \times \left(n_{L}\Upsilon\right)}$ is a selection matrix. The multi-layer perceptron (MLP) for the $\upsilon^{th}$ branch, $\text{MLP}_{\upsilon}(h)$, is given as
\begin{subequations}
% \vspace{-0.3cm}
\begin{equation}
\text{MLP}_{\upsilon}(h) :=W_{L\upsilon}\,\phi_{(L-1)\upsilon}\left(\dots\phi_{0\upsilon}(h)\right) + b_{L\upsilon},
% \vspace{-0.2cm}
\end{equation}
where 
% \vspace{-0.3cm}
\begin{equation}
\qquad\phi_{l\upsilon}(h)=\sigma\left(W_{l\upsilon}h+b_{l\upsilon}\right), \quad l=0,\dots,L-1,
% \vspace{-0.2cm}
\end{equation}
\end{subequations}
and the weight matrix and bias vector are given as $W_{l\upsilon} \in \mathbb{R}^{n_{l+1} \times n_l}$ and $ b_{l\upsilon} \in \mathbb{R}^{n_{l+1}}$. The hyperbolic tangent is used as the activation function $\sigma$. Collectively, the parameters of the neural ODE are denoted by $\theta := \left\{\left(W_{01},\,b_{01}\right),\,\dots,\,\left(W_{L\Upsilon},\,b_{L\Upsilon}\right)\right\}$.

This architectural choice in \eqref{eq:neural_ode} imposes structural knowledge about the disturbances experienced by the bottom quadrotor. First, choosing the inputs as in \eqref{eq:h0} results in a neural ODE that is axially symmetric about the $z$-axis of $\calF$. Second, the MLP outputs in \eqref{eq:mlp_output} are enforced to be in $\calF$ by scaling and rotating them into the required reference frames, outside of the network. This is done by multiplying them by the matrix $H$, followed by $M \in \mathbb{R}^{13\times13}$ given by
% \vspace{-0.1cm}
\begin{subequations}
\begin{equation} \label{eq:m_matrix}
M := \begin{bmatrix} M_{11} & 0_{10\times 3}\\
    0_{3 \times 10}& \inv{\hat{J}} R^{\top} R_{\calW}^{\calF}{}^{\top} 
    \end{bmatrix},
\end{equation}
where
% \vspace{-0.1cm}
\begin{equation}
    M_{11} = \begin{bmatrix}  & 0_{3\times 10} & \\
    0_{3\times 3} & \tfrac{R_{\calW}^{\calF}{}^{\top}}{m} & 0_{3\times 4} \\
     & 0_{4\times 10} & 
    \end{bmatrix},
% \vspace{-0.1cm}
\end{equation}
\end{subequations}
and $\inv{\hat{J}} := \inv{J} / ||\inv{J}||_F$ where $||\cdot||_F$ is the Frobenius norm.

Combining the neural ODE with the knowledge described in Section \ref{subsec:dynamics}, the overall KNODE-DW model is given as
% \vspace{-0.15cm}
\begin{equation} \label{eq:knode_model}
    \dot{x}(\theta) = f_{nom}(x,u) + f_d(x, x_{top}) + d\left(x,x_{top}\, ; \theta\right),
% \vspace{-0.15cm}
\end{equation}
and we denote a model consisting of only the components $f_{nom}(x,u)$ and $d\left(x,x_{top} \,; \theta\right)$ as the KNODE model. We highlight that our approach is fundamentally different from those in the existing literature, \textit{e.g.}, \cite{li2023nonlinear, shi2021neural}. We leverage analytical models and use a data-driven component to capture the residual aerodynamics, while existing works use the data-driven component to capture the full aerodynamics, similar to a KNODE model.

To train the model, we follow a procedure that is similar to those described in \cite{chee2022knode, jiahao2022online}. Given a dataset $\calD = \left\{\left(x_i,\, x_{top,i},\, u_i\right)\right\}_{i=1}^M$, we define the loss function to be a weighted mean squared error (MSE) between the one-step state predictions and the true states,
% \vspace{-0.2cm}
\begin{equation}
    \mathcal{L}(\theta) := \frac{1}{M-1}\sum^{M}_{i=2}\big\|\big(\hat{x}_i(\theta) -  x_i\big)\big\|^2_{W_x},
    \label{eq:loss}
% \vspace{-0.1cm}
\end{equation}
where $W_x \in \mathbb{R}^{n \times n}$ is a user-specified weighting matrix. The one-step predictions are given as 
% \vspace{-0.1cm}
\begin{equation}
    \hat{x}_{i}(\theta) := x_{i-1} + \int^{t_{i}}_{t_{i-1}} \dot{x}_{i-1}(\theta)\, dt, \quad i=2,\dots,M
% \vspace{-0.1cm}
\end{equation}
where %$\dot{x}(\theta)$ is derived from \eqref{eq:knode_model} and 
the integration is done using a given numerical solver. Next, we leverage deep learning tools, \textit{e.g.}, \texttt{PyTorch} \cite{paszke2017automatic}, to find a set of optimal parameters $\theta^{\star}$. In particular, we compute the gradient of the loss function $\mathcal{L}(\theta)$ through a backpropagation procedure, and incorporate them into an optimization algorithm, \textit{e.g.}, Adam \cite{Kingma2015AdamAM}, which solves for the parameters $\theta^{\star}$ in an iterative manner. 

One advantage of neural ODEs, as compared to standard networks, is that force and torque measurements are not required. In practice, the acceleration measurements from the inertial measurement unit (IMU) are noisy and may contain non-constant biases. Furthermore, the IMU only provides measurements of the angular velocities. This implies that the moments of inertia have to be sufficiently accurate such that accurate torque estimates can be extracted. On the other hand, velocity and orientation estimates are more easily attainable and are often more accurate.

% \vspace{-0.15cm}
\subsection{Incorporating into Learning-based MPC} \label{subsec: integrate_lmpc}
% \vspace{-0.1cm}
After training, we incorporate the learned KNODE-DW model into the optimization problem by replacing \eqref{eq:dynamics_const} with an updated set of dynamics constraints. These constraints are constructed by discretizing \eqref{eq:knode_model} with the parameters $\theta^{\star}$ using a suitable numerical method, \textit{e.g.}, 4$^{\text{th}}$ order Runge Kutta. At each time step $k$, the constraints are given as,
% \vspace{-0.15cm}
\begin{equation}
    x_{i+1} := \hat{f}(x_i,u_i,x_{top,i}(k); \,\theta^{\star}), \quad i=0,\dots,N-1.
% \vspace{-0.12cm}
\end{equation}
To incorporate the states of the top quadrotor, we formulate \eqref{eq:ftocp} as a parametric optimization problem. In this formulation, the current state $x(k)$, the reference states $\{x_{r,i}(k)\}_{i=0}^{N}$ and the states of the top quadrotor $\{x_{top,i}(k)\}_{i=0}^{N-1}$ act as parameters to the problem. To obtain the states of the top quadrotor along the prediction horizon, we assume that it travels at a constant velocity along the prediction horizon, and propagate its position using this velocity.  

Using an optimization-based control formulation like MPC provides two advantages. First, given a sequence of position estimates of the top quadrotor, the bottom quadrotor is able to anticipate and account for the aerodynamic disturbances in the next few timesteps of its flight. In particular, the problem \eqref{eq:ftocp} considers the proximity of the top quadrotor when computing the control actions for the bottom quadrotor. This is in stark contrast with some of the current control architectures for close proximity flight, \textit{e.g.}, \cite{smith2023so, shi2020neural}. These controllers only considers the disturbances caused by the top quadrotor at the current time step. Second, since it is an optimization-based scheme, it is straightforward to include collision avoidance constraints into \eqref{eq:ftocp}, given the position of the quadrotors. We leave the integration of these constraints as part of future work.

The incorporation of the KNODE-DW model into the MPC framework is a natural, yet significant, extension of our prior work \cite{chee2022knode}. The key architectural difference is the addition of the disturbance model \eqref{eq:f_d}, during training and deployment. While our proposed approach is conceptually similar, it is shown in our experiments that the framework in \cite{chee2022knode} does not necessarily lead to accurate closed-loop performance under all test cases. %, especially those involving stronger aerodynamic disturbances from the top quadrotor.
A detailed discussion is given in Section \ref{sec:expt_results}.

% \vspace{-0.1cm}
\section{EXPERIMENTS AND RESULTS} \label{sec:expt_results}
% \vspace{-0.1cm}
In our experiments, we set out to answer the following questions about our proposed framework: (i) Does the KNODE-DW model provide accurate predictions of the disturbance forces and torques caused by the top quadrotor? (ii) By incorporating the model into learning-based MPC, are there significant improvements in terms of closed-loop trajectory tracking performance, over meaningful baselines? (iii) Do the model and control framework generalize to cases beyond in which training data is collected? (iv) Do the results observed in simulations extend to physical experiments? (v) Can the quadrotors fly in a tight formation with the proposed framework?

We consider a number of baselines to ascertain the efficacy of our framework. First, we consider a nominal MPC framework, which consists of the nominal dynamics $f_{nom}$. This allows us to examine the improvements brought forth by the disturbance model $f_d$ and the neural ODE $d(\theta^{\star})$. Next, we consider a KNODE-MPC framework, which leverages $f_{nom}$ as knowledge, and uses a neural ODE to account for all unmodeled dynamics, including the aerodynamic disturbances. Third, we consider DW-MPC, which only comprises of $f_{nom}$ and $f_d$ and there are no learning components in it. In the simulations, we consider an additional benchmark, an omniscient MPC framework, where the model is identical to the true system dynamics. Finally, we have our proposed framework, KNODE-DW MPC, which not only consists of both $f_{nom}$ and $f_d$, but also includes a neural ODE $d(\theta^{\star})$.

% \vspace{-0.2cm}
\subsection{Simulations} \label{subsec:sim}
% \vspace{-0.15cm}
\textit{Setup:} A simulator is constructed using \eqref{eq:quad_dyn} and the disturbance model \eqref{eq:f_d}. An explicit 5$^{\text{th}}$ order RK method (RK45) with a time step of 5 ms is used for numerical integration to simulate the responses for the system. The disturbances \eqref{eq:aero_force_torque} are implemented using a first-order approximation of the 2D trapezoidal rule. To ascertain the efficacy of the neural ODE, we consider a mismatch between the disturbance model in the simulator and that in the controller. The simulated values of $C_D$ and $S$ are 1.3 and 0.0697, and these values in the controller are 1.18 and 0.0997. 

The learning-based MPC framework is assumed to have access to the states of both quadrotors. The framework is implemented using \texttt{acados} \cite{Verschueren2021} and \texttt{CasADi} \cite{Andersson2019}. It generates thrust and moment commands, which act as inputs to the simulator which generates the closed-loop responses. In simulations, we consider a two-branch, 1-layer neural ODE with 9 neurons for each branch. The elements of the selection matrix $H_{ij}$ are set to be 1 for \small$(i,j) \in \{(4,4),(5,5),(6,6),(4,17),(5,18),(6,19)\}\,$\normalsize and 0 elsewhere, and the two branches represent the residual disturbance forces and torques. 

For two quadrotors moving in a straight path, we observe that the case of crossing trajectories, \textit{i.e.}, when their 2D trajectories overlap momentarily, is significantly more benign, as compared to the cases of a static top quadrotor or when the quadrotors are in a stacked formation. This is because the time period in which the bottom quadrotor experiences the disturbances is shorter. While the case of crossing trajectories is more commonly considered in the literature, \textit{e.g.}, \cite{smith2023so, li2023nonlinear}, we are interested in testing the limits of our framework. That motivates us to consider these two latter cases, which we denote as \emph{static top} and \emph{stacked}, in both simulations and physical experiments. 

The training data $\calD$ consists of four flight segments. Two of the segments are from the \emph{static top} case, while the other two are from the \emph{stacked} case. The dataset consists of 3,532 data points, which corresponds to approximately 18 seconds of flight. They are obtained from closed-loop simulations with nominal MPC, at a speed of 0.4 m/s and separations of 0.3 m and 0.4 m. We set $W_x := \text{diag}\left(\mathbb{I}_{10 \times 10},\,\,0.1\, \mathbb{I}_{3 \times 3}\right)$ during training. Apart from training cases, additional tests are conducted at two other speeds, 0.3 m/s and 0.5 m/s, and at two other separations of 0.35 m and 0.2 m. This is to evaluate the generalization abilities of the model and controller.

\begin{figure}
    \centering
    {\includegraphics[scale=0.22, trim = 0cm 0.7cm 0cm -0.9cm]{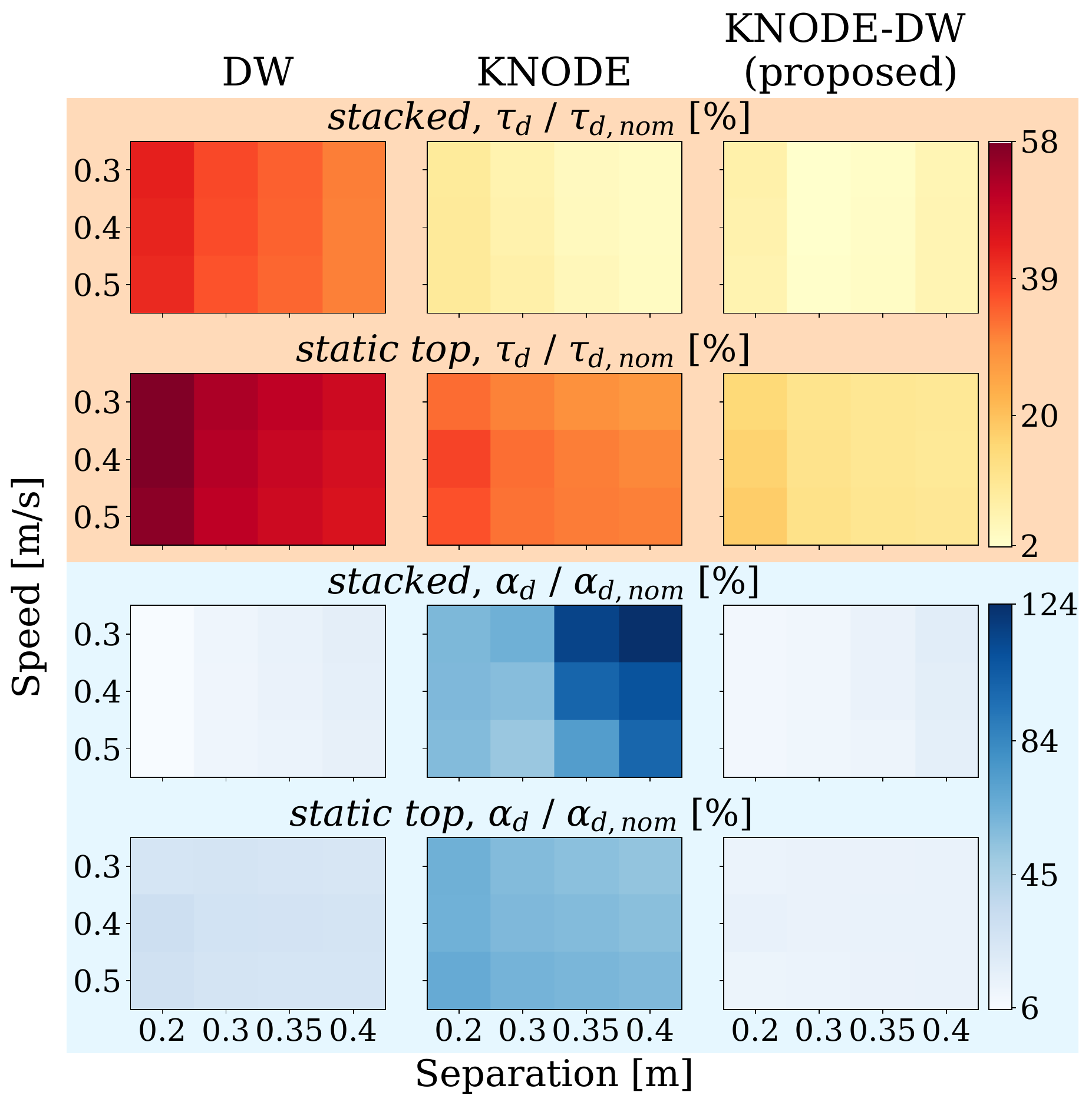}}
    \caption{\small\textbf{Force $(\tau_d)$ and torque $(\alpha_d)$ predictions:} Heatmaps of the prediction root mean squared errors (RMSEs) given by the KNODE, DW and KNODE-DW models, normalized against those given by the nominal model, under different speeds and vertical separations. %The first and third rows are for the \textit{stacked} formation, and the second and fourth rows are for the case of a \textit{static top} quadrotor.
    }
    \label{fig:sim_predictions}
\end{figure}

\begin{figure}
    \centering
    {\includegraphics[scale=0.21, trim = 1cm 0.7cm 0cm -0.9cm]{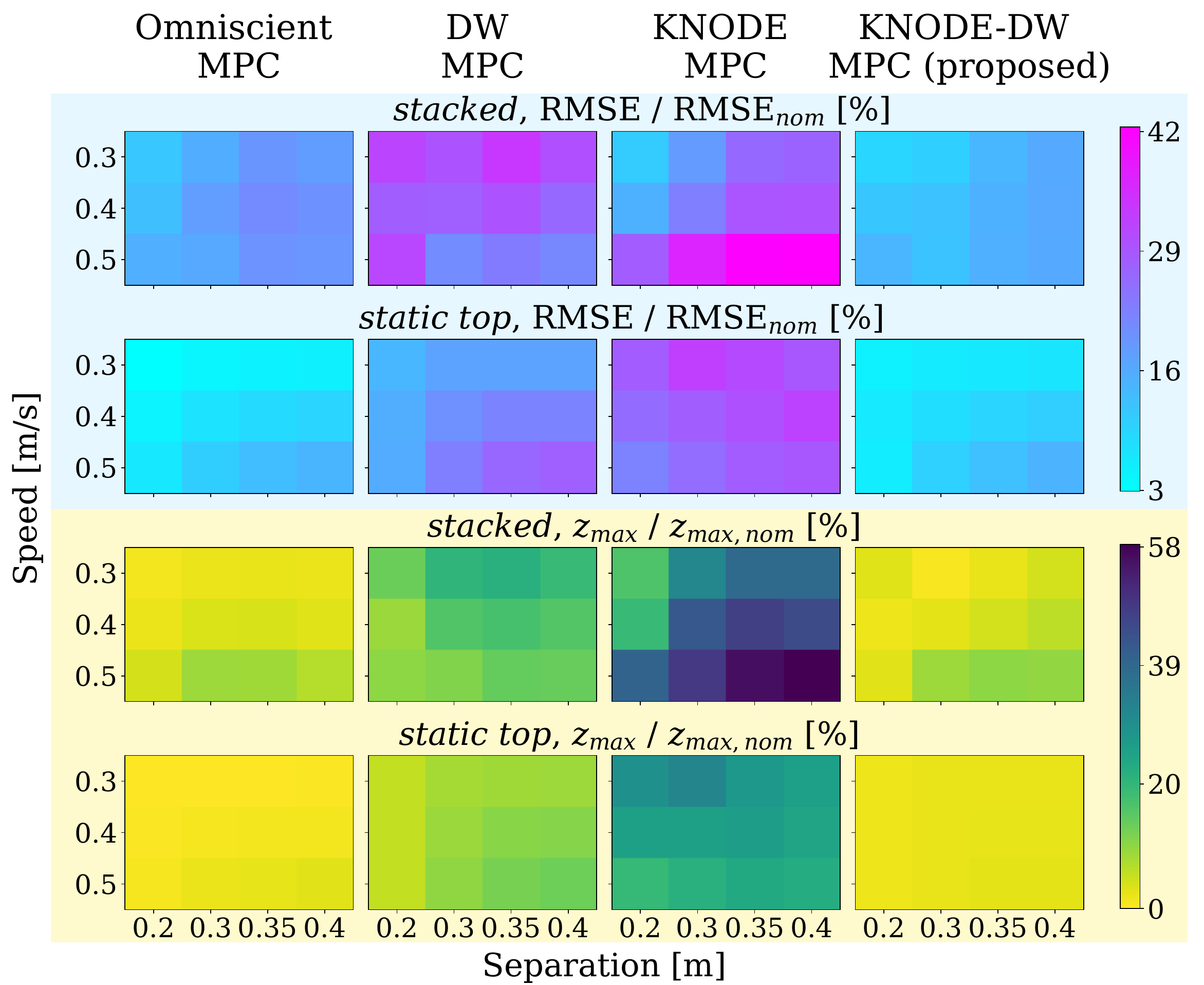}}
    \caption{\small\textbf{Closed-loop performance:} Heatmaps of the tracking RMSEs and maximum vertical separation $\left(z_{{max}}\right)$ of the baseline and proposed MPC frameworks, normalized against those obtained from nominal MPC, under different speeds and vertical separations. %The first and third rows are for the \textit{stacked} formation, and the second and fourth rows are for the case of a \textit{static top} quadrotor.
    }
    \label{fig:sim_rmse}
\end{figure}

\textit{Results:} To verify the accuracy of the models, we compare the predictions of the disturbance forces and torques against with the true disturbances. As seen from Fig. \ref{fig:sim_predictions}, the KNODE-DW model provides the most accurate predictions in both the \emph{stacked} and \emph{static top} cases, attaining a 89.5\% improvement over the nominal model, on average. While the KNODE model achieves similar improvements in force predictions under the \emph{stacked} formation, the prediction errors are relatively large under the \emph{static top} scenario. Furthermore, the KNODE model does not yield accurate predictions for the torques, performing worse than the nominal model at times. In contrast, the DW model provides more consistent predictions for the torque predictions, but fares worse for the force predictions. Overall, the KNODE-DW model combines the strengths of both the KNODE and DW models. We highlight that in the current literature, disturbance torques are typically not considered within the controllers, \textit{e.g.}, \cite{smith2023so, li2023nonlinear, shi2020neural}. In contrast, our proposed framework predicts and compensates these torques in a systematic manner, as shown in Figs. \ref{fig:sim_predictions} and \ref{fig:sim_rmse}.

Next, we compare the proposed KNODE-DW MPC framework against the baselines, in terms of the closed-loop tracking performance. We consider two metrics, the root mean squared errors (RMSEs) between the bottom quadrotor and reference trajectories, and the maximum vertical separation $z_{max}$, which is computed by taking the longest distance between the bottom quadrotor and reference heights.
It is observed in Fig. \ref{fig:sim_rmse} that all four control frameworks perform better than nominal MPC, since all the percentages are less than 100\%. Notably, KNODE-DW MPC achieves performance that is comparable to omniscient MPC in all cases, reducing the average RMSEs and $z_{max}$ by 89\% and 96.5\% over nominal MPC. It is observed that KNODE-DW MPC and DW MPC perform better than KNODE MPC on average, and DW MPC is consistent than KNODE MPC across the test cases. This illustrates the importance of embedding knowledge into training and within the control framework. 

% \vspace{-0.2cm}
\subsection{Physical Experiments} \label{subsec:expt}
% \vspace{-0.1cm}
\textit{Setup:} Two Crazyflie 2.1 quadrotors \cite{Bitcraze} are deployed in the experiments. Each of the quadrotors has a body length of 10 cm and weighs approximately 34g. The bottom quadrotor is equipped with the thrust upgrade bundle \cite{Bitcraze} to increase its control authority. A laptop running on Intel i5 CPU acts as the base station and communication between the Crazyflies are established via Crazyradio PA at an average rate of 400 Hz. A Vicon motion capture system is used to obtain pose measurements and communicates with the base station at an approximate rate of 120Hz. The CrazyROS wrapper \cite{crazyflieROS} is used as part of the software architecture.

The learning-based MPC framework is implemented using the same software as that in simulations and operates on the base station. Instead of generating thrust and moment commands, the framework in physical experiments generates thrust and angular rate commands, which are passed to the PID controllers running in the Crazyflie firmware. With this control architecture, we compensate for the disturbance forces and not the torques, by setting $M := M_{11}$. We define $d(x,x_{top}; \theta)$ to be a single branch, single layer neural ODE with $4$ hidden neurons. The elements of the selection matrix $H_{ij}$ are 1 for \small$(i, j) \in \{(4, 4), (5, 5), (6, 6)\}\,$\normalsize and 0 elsewhere. To obtain an approximate model for the disturbance forces, we conduct a system identification procedure where we adjust the model parameters so that DW MPC attains reasonable performance under training conditions, and $C_D$ is set at 0.236.

Similar to the simulations, the training data is collected under the \textit{static top} and \textit{stacked} cases at a speed of 0.4m/s and a separation of 0.4m. The dataset consists of 6,720 data points, corresponding to 28 seconds of flight. This is done in closed-loop by flying the quadrotors under nominal MPC. This is in stark contrast with \cite{li2023nonlinear}, where the model is trained using 570 seconds of data. To ascertain generalization, another test case at a speed of 0.4m/s and at a smaller separation of 0.2 m is considered.

\begin{figure}
    \centering
    {\includegraphics[scale=0.245, trim = -0.2cm 0.5cm 0cm -0.9cm]{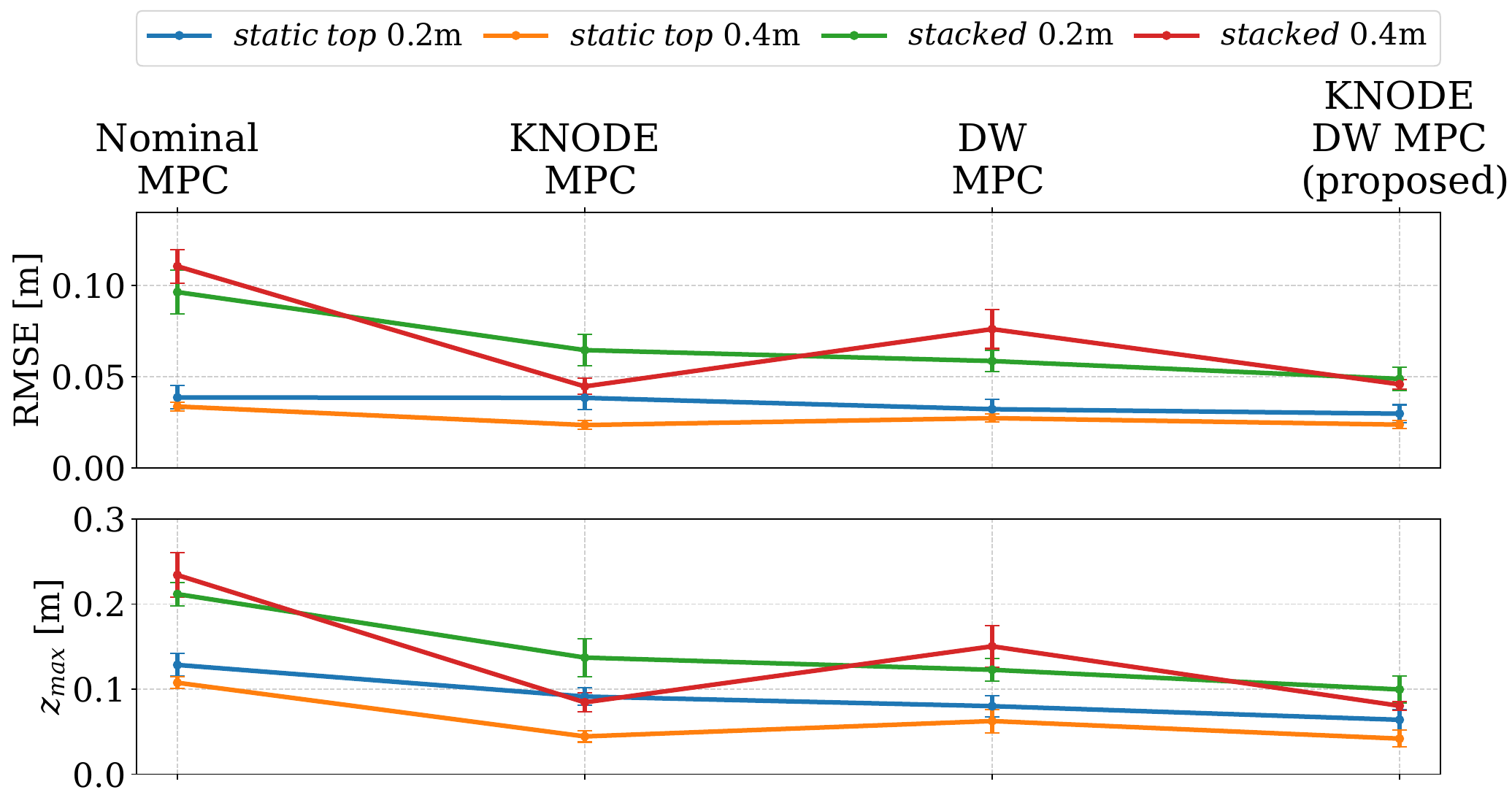}}
    \caption{\small\textbf{Experiment statistics:} Statistics of the runs for the baselines and the proposed framework. The top subplot depicts the RMSEs, and the maximum vertical separation $(z_{max})$ is shown in the bottom subplot. The markers and the error bars indicate the mean and standard deviation of the runs. The values behind the test cases in the legend, \textit{e.g.}, 0.2m for \textit{stacked} 0.2m, denote the commanded vertical separation between the two quadrotors.}
    \label{fig:phy_expt_stats}
\end{figure}

\begin{figure}
    \centering
    {\includegraphics[scale=0.25, trim = -0.2cm 0.5cm 0cm 0cm]{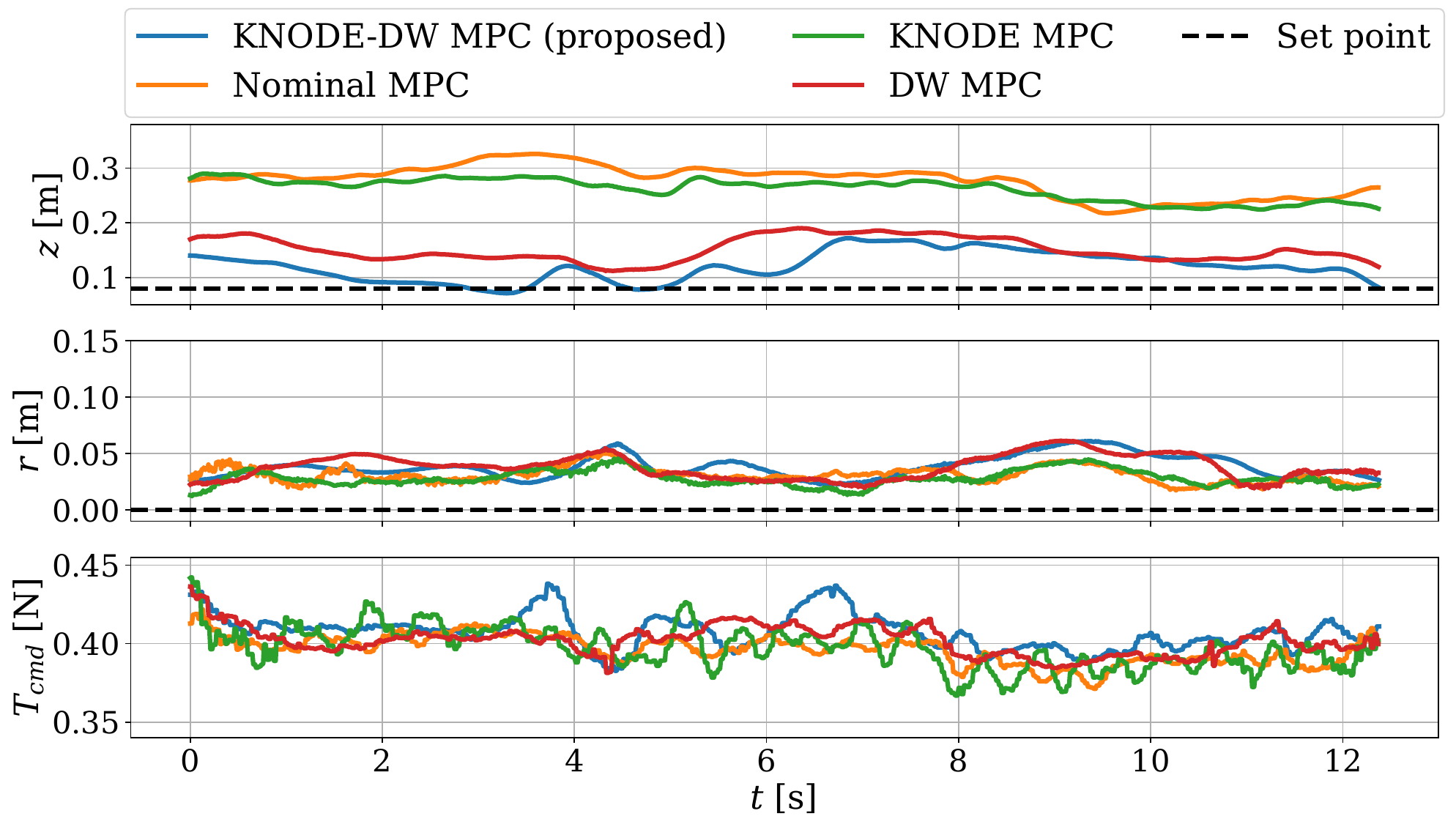}}
    \caption{\small\textbf{Experiment results:} Time histories of the vertical $(z)$ and radial $(r)$ separations, and the commanded thrust $(T_{cmd})$ of the bottom quadrotor, under the baselines and KNODE-DW MPC. This is during a more challenging test case, where the vertical separation between the two quadrotors is set at 0.8 body lengths throughout the flight.}
    \label{fig:phy_expt_time}
\end{figure}

%For the test case of swapping trajectories, in addition to the shorter time period as described in Section \ref{subsec:sim}, we hypothesize that there is a change in the velocity flow field below the top quadrotor when it travels with a non-zero velocity. This causes an even more benign disturbance to the bottom quadrotor in physical experiments.

\textit{Results:} Fig. \ref{fig:phy_expt_stats} depicts the statistics of the runs under different baselines. For each test case, 5 runs are conducted. First, we observe that all 3 baselines, KNODE MPC, DW MPC and KNODE-DW MPC, perform better than nominal MPC, which demonstrate the importance of having an accurate model in the controller. Moreover, it is observed that the KNODE-DW MPC framework not only achieves the lowest RMSEs, but also maintains the smallest $z_{max}$ across all test cases. In particular, our proposed framework achieves an average of 40.1\% improvement in RMSE and 57.5\% reduction in $z_{max}$ over nominal MPC across the test cases. While KNODE MPC provides comparable performance in cases where training data is collected, it fails to provide consistent performance for the other two test cases, unlike KNODE-DW MPC. The improvement of KNODE-DW MPC over DW MPC indicates that the neural ODE is able to compensate for the mismatch between the knowledge component and the true system dynamics.

To further evaluate the performance of our framework, we first consider a complex flight formation, as shown in the composite photo in Fig. \ref{fig:cover_image}. The two CFs are flying in vertical lemniscate trajectories, while maintaining a stacked formation. Even though the model has not seen this trajectory during training, it performs remarkably well, highlighting the generalization ability of our framework. Next, we consider a more challenging \emph{stacked} formation where the two quadrotors fly in a straight line, under a \emph{very} tight formation. They are commanded to be 0.8 body lengths (0.08m) vertically apart, throughout the flight. The time histories of the vertical separation $z$ is shown in Fig. \ref{fig:phy_expt_time}. Under KNODE-DW MPC, the average vertical separation between the drones is 0.12m, which is 57.1\% smaller than that when nominal MPC is applied, which is at 0.28m. We highlight that this vertical separation of 1.2 body lengths is significantly smaller than what has been reported in the literature, which is of at least 2 body lengths, \textit{e.g.}, \cite{shankar2023docking, shi2020neural}. We also note that the average radial separation $r$ under our framework is at 0.03m, which implies that the bottom quadrotor is well in the wake of the top quadrotor. The difficulty of flying in such a tight formation stems from the complex aerodynamic disturbances experienced by the bottom quadrotor, when it is within 2 body lengths of the top quadrotor \cite{bauersfeld2024robotics, kiran2024downwash}.
We demonstrate with this demanding test case that our framework is able to overcome these complex aerodynamic disturbances, enabling tight formation flight for the first time. Interested readers are referred to the accompanying video for more details on the physical experiments.

% \vspace{-0.2cm}
\section{CONCLUSION} \label{sec:conclusion}
% \vspace{-0.15cm}
In this work, we present a framework to model and learn the aerodynamic disturbances caused by another quadrotor in an accurate and sample efficient manner. It is demonstrated through simulations and physical experiments that the proposed model and control framework are highly accurate in terms of disturbance predictions, as well as closed-loop trajectory tracking performance. One possible direction of future work is to consider contemporary uncertainty quantification methods such as those in \cite{chee2024uncertainty}, to improve the uncertainty awareness in these tight formations.

% \vspace{-0.1cm}
\section{ACKNOWLEDGMENTS}
% \vspace{-0.1cm}
The authors thank Anthony Cowley for inspiring the photo in Fig. \ref{fig:cover_image}, and Pratik Kunapuli for providing CF components during part of the development.

\addtolength{\textheight}{-7cm}   % This command serves to balance the column lengths
                                  % on the last page of the document manually. It shortens
                                  % the textheight of the last page by a suitable amount.
                                  % This command does not take effect until the next page
                                  % so it should come on the page before the last. Make
                                  % sure that you do not shorten the textheight too much.

\bibliographystyle{IEEEtran} 
\bibliography{IEEEabrv,refs}

\end{document}